\documentclass{article}

\usepackage{arxiv}

\usepackage[utf8]{inputenc} 
\usepackage[T1]{fontenc}    
\usepackage{hyperref}       
\usepackage{url}            
\usepackage{booktabs}       
\usepackage{amsfonts}       
\usepackage{nicefrac}       
\usepackage{microtype}      
\usepackage{lipsum}
\usepackage{graphicx}
\usepackage{epsfig}
\usepackage{subcaption}
\usepackage{calc}
\usepackage{amssymb}
\usepackage{amstext}
\usepackage{amsmath}
\usepackage{amsthm}
\usepackage{multicol}
\usepackage{algorithm2e}
\usepackage[bottom]{footmisc}
\graphicspath{ {./images/} }

\title{Time-Constrained Recommendations: Reinforcement Learning Strategies for E-Commerce}

\author{
 Sayak Chakrabarty $^\dagger$ \\
  Department of Computer Science\\
  Northwestern University\\
  Evanston, IL 60208, USA\\
  \texttt{sayakchakrabarty2025@u.northwestern.edu} \\
   \And
 Souradip Pal $^\dagger$ \\
  Elmore Family School of Electrical and Computer Engineering\\
  Purdue University\\
  West Lafayette, IN 47906, USA\\
  \texttt{pal43@purdue.edu}
}

\begin{document}
\maketitle
\def\thefootnote{\dag}\footnotetext{These authors contributed equally to this work.}
\def\thefootnote{\arabic{footnote}}
\begin{abstract}
Unlike traditional recommendation tasks, finite user time budgets introduce a critical resource constraint, requiring the recommender system to balance item relevance and evaluation cost. For example, in a mobile shopping interface, users interact with recommendations by scrolling, where each scroll triggers a list of items called slate. Users incur an evaluation cost — time spent assessing item features before deciding to click. Highly relevant items having higher evaluation costs may not fit within the user’s time budget, affecting engagement. In this position paper, our objective is to evaluate reinforcement learning algorithms that learn patterns in user preferences and time budgets simultaneously, crafting recommendations with higher engagement potential under resource constraints. Our experiments explore the use of reinforcement learning to recommend items for users using Alibaba's Personalized Re-ranking dataset supporting slate optimization in e-commerce contexts. Our contributions include (i) a unified formulation of \emph{time-constrained slate recommendation} modeled as Markov Decision Processes (MDPs) with budget-aware utilities; (ii) a simulation framework to study policy behavior on re-ranking data; and (iii) empirical evidence that on-policy and off-policy control can improve performance under tight time budgets than traditional contextual bandit-based methods.
\end{abstract}


\section{Introduction}
\label{sec:introduction}

One key goal of modern recommender systems is to effectively enhance user engagement. In e-commerce, streaming, and news feeds, users interact with \emph{slates}—lists of $K$ items shown together—while constrained by a finite \emph{time budget}. Each item requires cognitive effort and time to assess before any click or purchase decision, like scroll or search, where users look through and evaluate items sequentially and may abandon once their time budget is exhausted. Conventional ranking algorithms that optimize only predicted relevance, often over-allocate attention to costly items and under-perform when users stop early. This paper argues that \emph{time} is a crucial resource in such recommendation scenarios and that algorithms must explicitly trade off \emph{utility} (engagement, conversion) with \emph{evaluation cost} (time-to-assess), at the slate level. Time-aware slates can front-load high utility-per-second items to reduce abandonment. For example, per-item skip costs vary for short-video feeds; optimizing clicks or watch-through under a per-session time budget improves session satisfaction. Long-form articles or lectures in news and learning apps have high evaluation costs; slates should mix quick wins with deeper content tied to user intent. Also, ads and sponsored content requires pacing under attention budgets which can raise total utility while respecting platform constraints.

\begin{figure}[ht]
\centering
  \includegraphics[width=0.8\linewidth]{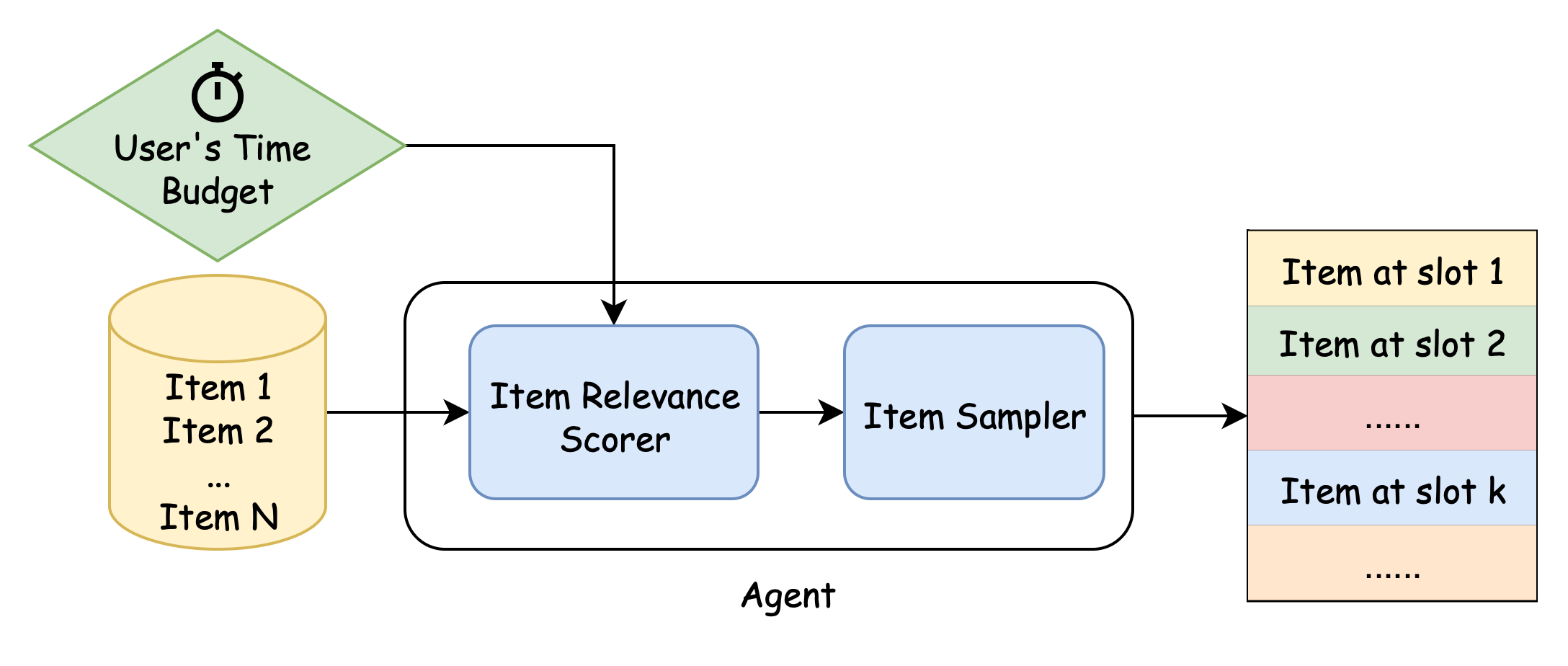}
  \caption{Slate Recommender System}
  \label{fig:slate}
\end{figure}

The main aim of slate recommenders is to insert the most relevant element at slot $k$ in the slate where an item relevance scorer generates relevance scores of the available $N$ items and using the slate constructed so far as additional context. The scores are then passed through a sampler to select an item from the available items as shown in Figure \ref{fig:slate}. In this \emph{position paper}, we evaluate \emph{time-aware slate optimization} using reinforcement learning (RL) techniques and outline open research problems. We also provide initial evidence using Alibaba's Personalized Re-ranking (PRM) dataset \cite{alibabaPRMdataset,pei2019personalized}, where we simulate user budgets and evaluation costs and compare on-policy (SARSA) \cite{Rummery1994OnlineQU,zhang2022new,afsar2022reinforcement,zhang2023sockdef} and off-policy (Q-Learning) \cite{Watkins1992-ni,NIPS2017_5352696a,Chaudhari_Arbour_Theocharous_Vlassis_2024} controllers. Apart from using ID-based features as done in case of Personalized Re-ranking Model (PRM), other models like LLMs or VLMs or techniques like logic-based programming and bootstrapping methods \cite{bolonkin2024judicial,chakrabarty2024mm,chakrabarty2025readmeready,banerjee2025clt}, also provide suitable ways to extract text/image based item features allowing better ranking of items in a slate. The remainder of the paper reviews related work on slate reinforcement learning and budget-aware recommendation, details our simulation approach, reports comparative results for SARSA and Q-Learning, and discusses implications and open problems for the future.

\section{Related Works}

Traditional recommender systems typically score and rank items independently, but in many applications, the system must recommend a slate of multiple items at once. This introduces new challenges since user interactions with a slate can exhibit position biases and inter-item effects. Early work by \cite{sunehag2015deep} formalized the Slate MDP (Markov Decision Process) concept to model recommendation lists as combinatorial actions. In a Slate MDP, the agent suggests a tuple of items, and the user may engage with one (or none), which complicates the reward attribution. The large action space of possible slates makes direct RL difficult, so subsequent approaches have sought tractable approximations. For example, SlateQ \cite{ie2019slateq} decomposed the long-term value of a slate into the Q-values of individual items. Under assumptions like at most one item per slate is consumed, this decomposition allows standard Q-Learning updates at the item level. Similar item-level ranking strategies have been used by others \cite{deffayet2023generative,zhao2017deep,zhao2018deep,lin2023survey} to construct slates by sampling items from a learned scoring policy, albeit with simplifying independence assumptions. More recent work has tried to relax these assumptions. For instance, \cite{deffayet2023generative} proposes a generative slate recommendation approach using a VAE to encode entire slates into a continuous latent space, enabling an RL agent to generate diverse slates without evaluating all combinations. These developments highlight the progression from treating slate recommendation as a combinatorial bandit problem to a sequential decision problem tackled with RL techniques.

Reinforcement learning in \cite{elahi2020budget} has been applied to optimize long-term user engagement in recommendation scenarios \cite{zheng2018drn,ie2019reinforcementlearningslatebasedrecommender} (e.g. maximizing session duration or total clicks) \cite{10.1145/3292500.3330668}, for example, demonstrated that RL policies can outperform myopic recommenders by considering future rewards and feedback loops. However, relatively few works explicitly incorporate resource constraints such as a user's limited attention or time. Recently, industry researchers have emphasized the importance of accounting for finite user time budgets. In contrast to prior RL recommenders \cite{liu2018deep} that mostly optimize click-through or long-term value without explicit budgets, we focus on scenarios where each user has a finite time budget, and we assess the algorithms' ability to maximize engagement (play rate) while respecting this budget through appropriate slate construction.

\section{Methodology}

\subsection{Problem Formulation}
We consider a slate recommendation scenario in which a user interacts with a recommender system throughout an episode of  $T$ turns. At every turn $t \in \{1,..., T\}$, the system recommends a slate $a_t= (i^1_t, i^2_t,...,i^K_t)$ where $(i^j_t)_{1 \le j \le K}$ are items from the collection $\mathcal{I}$ and $K$ is the size of the slate. The user can
click on zero, one or several items in the slate and the resulting click vector $e_t =(e^1_t, e^2_t,...,e^K_t)$, $e^k_t\in \{0,1\}$ is returned to the recommendation system. At each step, a relevance scorer estimates per-item relevance $\sigma_k\in[0,1]$ for position $k$. Let $\beta_k=\sigma_k\prod_{m=1}^{k-1}(1-\sigma_m)$ denote the probability that the user selects the $k$-th item having skipped earlier ones (Abandon probability = $\prod_{m=1}^{k} (1-\sigma_m)$); $\sum_{k=1}^K\beta_k$ is the slate-level selection probability. Item $i$ carries an evaluation cost $c_i$ measured in seconds. The initial budget $u_0$ captures time the user is willing to invest in the session. After each examined item, the remaining budget updates based on engagement (e.g., click). User responses can exhibit position bias, inter-item effects, and the possibility of engaging with zero or one item from the slate \cite{sunehag2015deep,ie2019slateq}. Practical systems therefore model how likely each item is to be examined at its position and how items compete or complement one another on the same screen.

This problem parallels the 0/1 Knapsack problem, where item relevance represents utility, evaluation cost represents weight, and the user's time budget acts as the capacity. If $\beta_i$ and $c_i$ are the utility and cost of the $i^{th}$ item and $u$ is the user time budget, then the budget constrained recommendations can be formulated as follows:
\[
\max_{S} \sum_{i \in S} \beta_i \quad \text{such that} \quad \sum_{i \in S} c_i \leq u
\]
where $S$ is the optimal item subset. Budgets may be latent but can be inferred from scroll speed, abandonment, and historical patterns. Unlike static optimization problems, this setting involves dynamic user interactions, making it well-suited for reinforcement learning (RL). This naturally motivates a Markov Decision Process (MDP) treatment with state $s_t=(u_t,q_t)$, where $u_t$ is time-to-go and $q_t$ encodes items already shown; action $a_t$ is the next slate; and reward reflects observed engagement. Integrating RL into the recommender architecture effectively transitions from traditional solutions, enhancing engagement.

\subsection{Slate MDP}
Here, the time-constrained recommendation task has been modeled as a Markov Decision Process (MDP), with the recommender as the agent and the user's interaction state defined by the remaining time budget and examined items. User time budgets, often unobservable, can be estimated from historical data such as scroll behavior. We instantiate a Slate MDP \cite{sunehag2015deep,ie2019slateq} where actions are slates and rewards derive from an intra-slate relevance scorer \cite{pei2019personalized} based on a Personalized Re-ranking Model (PRM). The MDP $\mathcal{M} = (\mathcal{S},\mathcal{A},\mathcal{R},\Omega)$ is defined as follows:
\begin{itemize}
\item \textbf{State @ Step $t$} ($s_t$):  $(u_t, q_t) \in \mathcal{S}$ where $u_t$ is the user budget to go and $q_t$ is  list of first $k-1$ items in the slate examined
\item \textbf{Action @ Step $t$} ($a_t$): $i^k_t \in \mathcal{A}$ where $i^k_t$ is the item at slot $k$ in the slate with relevance $\sigma_{i^k_t}$ and cost $c_{i^k_t}$
\item \textbf{Reward @ Step $t$} ($r_t$): $r_t \sim \text{Bernoulli}(\beta_{i^k_t})$ if $c_{i^k_t} \leq u_t$, else $0$
\item \textbf{State @ Step $t+1$} ($s_{t+1}$): $(u_{t+1}, q_{t+1}) \in \mathcal{S}$ where $u_{t+1} = u_t - c_{i^k_t}r_t, \quad q_{t+1} = q_t \cup i^k_t$ \\
\end{itemize}

Using reinforcement learning techniques like SARSA or Q-Learning, the system learns a policy to optimize slate composition, balancing relevance and cost. We compare on-policy SARSA and off-policy Q-Learning controllers that condition on remaining time budget and slate prefix. Controllers learn to prioritize items with favorable \emph{utility-per-second} when time budgets are tight, and to exploit longer-tail items when time budgets are ample.

\subsection{Simulation}
Simulations are a powerful tool for studying time-constrained recommendation systems in e-commerce. In this context, we model user interactions with slates generated from a subset of Alibaba's Personalized Re-ranking dataset \cite{alibabaPRMdataset}. Instead of sampling relative log-relevance $\log \beta_i$ for each item independently from a pre-defined distribution, we use a Personalized Re-ranking (PRM) model \cite{pei2019personalized} as an intra-scorer to generate item relevance score $\sigma_i$ from which we derive relative log-relevance $\log \beta_i$ for calculating the reward in each step. The PRM model was trained on 8000 interactions of 150 users and $N=142{,}998$ unique items, with each interaction providing a list of $K=30$ items. Each of these interactions has both user (3 categorical) and item (5 categorical + 12 dense) features and labels corresponding to the user clicks. The validation set consisted of 1000 interactions. The hyperparameters used for training the PRM model are given in Table \ref{table:hyperparams}.

\begin{table}[ht]
\centering
\begin{tabular}{cc}
    \toprule
    \textbf{Hyperparameter} & \textbf{Value} \\
    \midrule
    \texttt{emb\_dim} & 64 \\
    \texttt{n\_layers} & 2 \\
    \texttt{n\_heads} & 1 \\
    \texttt{hidden\_dim} & 128 \\
    \texttt{dropout} & 0.1 \\
    \texttt{learning\_rate} & 0.001 \\
    \bottomrule
\end{tabular}
\caption{Hyperparameters used for training the PRM model as an Item Relevance Scorer}
\label{table:hyperparams}
\end{table}

The RL simulation process relies on several parameters as shown in Table \ref{table:simparams}, such as budget and cost distributions, which can be adjusted to create diverse experiments. For our experiments, we sample the cost parameters from a uniform distribution $c_i \sim \text{Uniform}(low, high)$ while varying the location parameter ($loc_u$) of the initial user budget distribution  $u_0 \sim \log - \mathcal{N}(loc_u, scale_u)$ to analyze performance under different budget constraints. We then compute which items are affordable given the user's current budget. The next user choice is modeled as multinomial logits over item relevances plus a ``no-choice" option where one observation (item or no-choice) is sampled per user ($o_t = Categorical(i_i, ..., i_k, \text{No-Choice})= Softmax (log \beta_1, ... , log \beta_k, log \beta_{No-Choice})$). The discounted return for each state, action pair is computed as $Q_\pi(s_t, i_t) = \mathbb{E}_\pi[\sum_{m=0}^{K}(\gamma^m r_{m+k})]$. This gives a single trajectory (episode) of interactions between the recommender, users, and items. The RL agent uses a gradient-boosted regression tree (XGBoost) \cite{10.1145/2939672.2939785} to approximate the $Q$-function and produces new slates for users by $\epsilon$-greedy exploration based on the SARSA or Q-Learning targets, following an $\epsilon$-greedy policy ($\pi$) to select the next best item. Our code is available at \url{https://github.com/souradipp76/Budget-Constrained-RLRS}.

\begin{table}[ht]
\centering
\begin{tabular}{cc}
    \toprule
    \textbf{Parameter} & \textbf{Value} \\
    \midrule
    \texttt{num\_users} & 150 \\
    \texttt{num\_items} ($N$) & 142998 \\
    \texttt{slate\_size} ($K$) & 30 \\
    \texttt{cost\_low} ($low$) & 0 \\
    \texttt{cost\_high} ($high$) & 100 \\
    \texttt{user\_budget} ($loc_u$) & \{$100, 150, ..., 500 $\} \\
    \texttt{user\_budget\_scale} ($scale_u$) & 0.5 \\
    \texttt{epsilon} ($\epsilon$) & 0.1 \\
    \texttt{discount\_factor} ($\gamma$) & \{$0.2, 0.4, ..., 1.0 $\}\\
    \bottomrule
\end{tabular}
\caption{Simulation parameters for SARSA and Q-Learning}
\label{table:simparams}
\end{table}

\begin{figure*}[ht]
\centering
  \includegraphics[width=\linewidth]{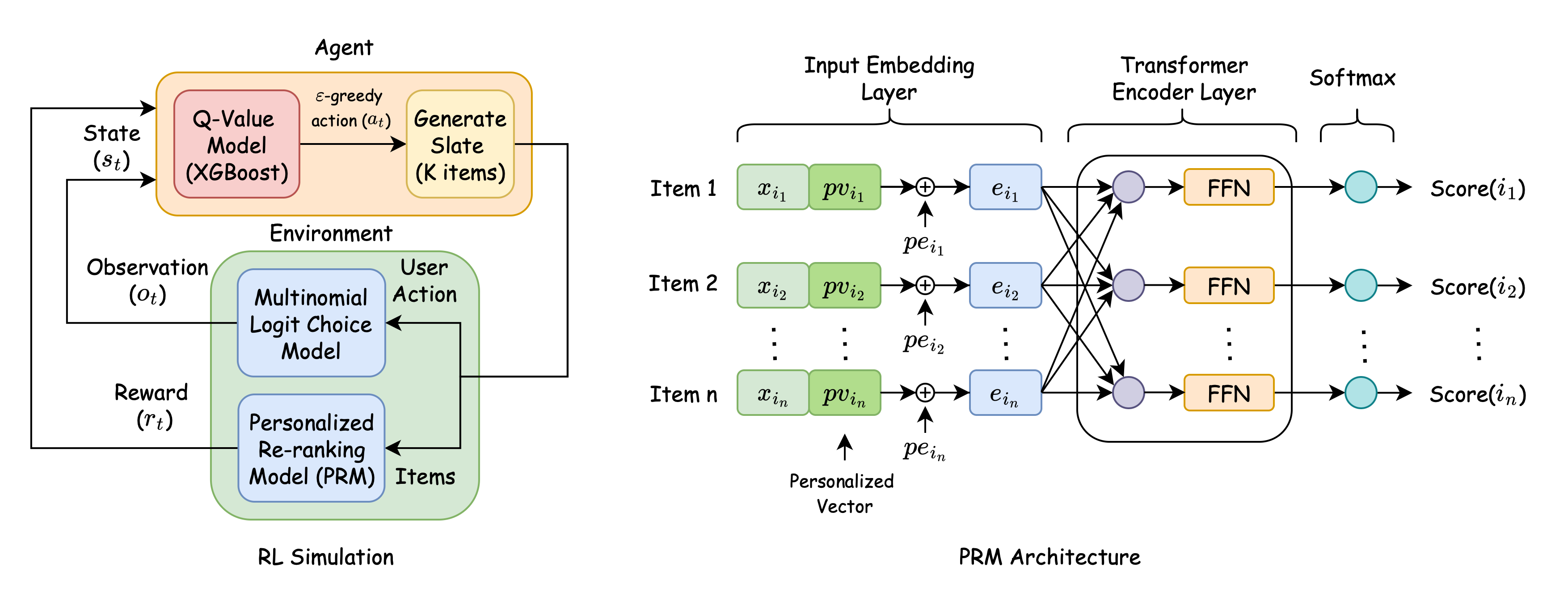}
  \caption{Overview of reinforcement learning simulation workflow for slate recommendation along with architecture of the PRM model for reward computation}
  \label{fig:prm}
\end{figure*}

\subsection{Evaluation Metrics}
Recommendation algorithms were evaluated using the following two key metrics.
\begin{itemize}
    \item \textbf{Play Rate} -- It measures the average number of successful engagements (clicks) from the generated slates, serving as a proxy for user engagement.
    \item \textbf{Effective Slate Size} -- Defined as the number of items in the slate that fit within the user's time budget before abandonment.
\end{itemize}

Maximizing play rate often involves constructing larger effective slates with relevant items, making this metric critical for understanding algorithm performance. It implies maximizing utility-per-second early in the slate while maintaining diversity and coverage. Together, these metrics provide insights into the ability of algorithms to balance relevance and cost under real-world budget constraints.

\section{Preliminary Results}

In simulations, Q-Learning maintains higher play rate and scales effective slate size with the discount factor, while SARSA benefits from larger budgets that average out exploratory variance. Both RL methods reduce abandonment relative to myopic contextual ranking, especially under small to medium time budgets. These patterns support our claim that explicit time budget conditioning at the slate level is beneficial.

\subsection{On-Policy Learning}
We trained recommendation policies in an on-policy manner to construct optimal slates under time budget constraints. This was achieved through the SARSA algorithm, where an initial random value function model iteratively updated based on user feedback until convergence. We evaluated the policies on metrics like play rate (average user engagement) and effective slate size (number of items fitting within the user’s time budget) for different user budgets.

\begin{figure*}[ht]
\centering
  \includegraphics[width=1\linewidth]{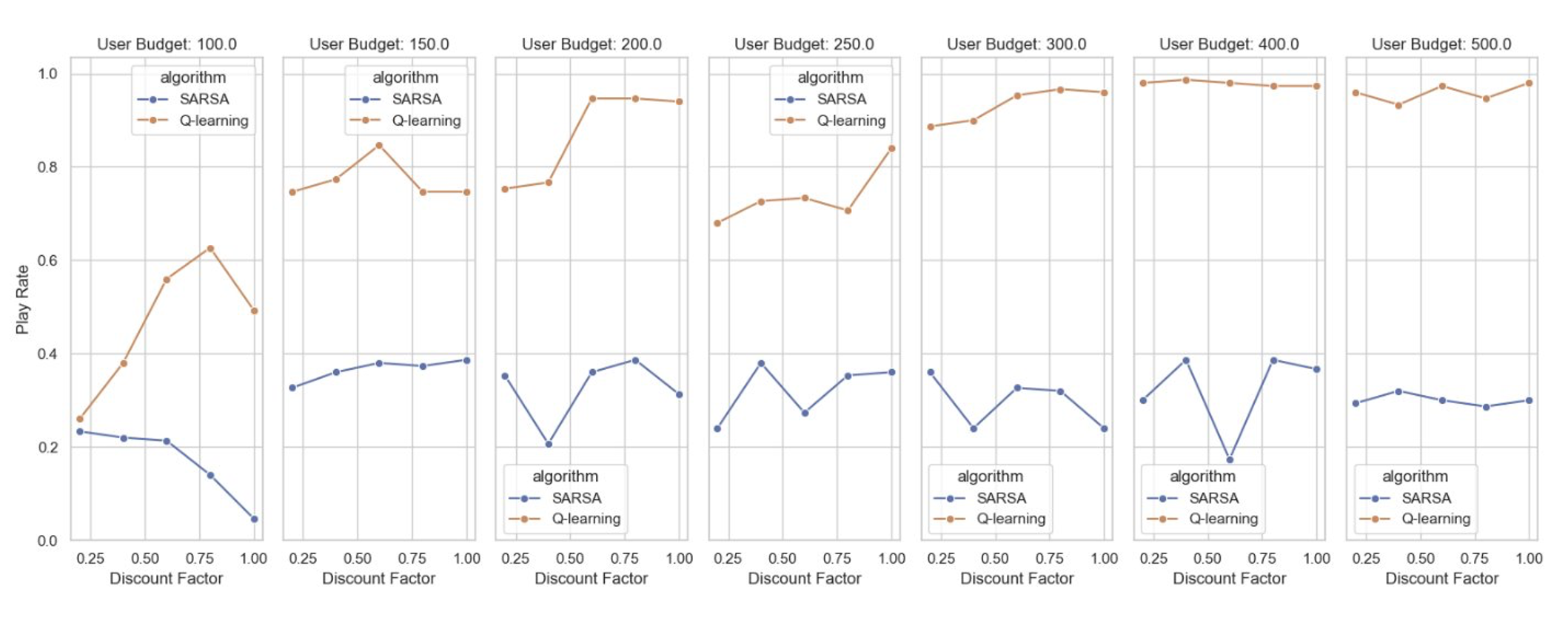}
  \caption{Plots showing variation of Play Rate with discount factor ($\gamma$) with fixed time budgets for SARSA and Q-Learning}
  \label{fig:play_rate}
\end{figure*}

\begin{figure*}[ht]
\centering
  \includegraphics[width=1\linewidth]{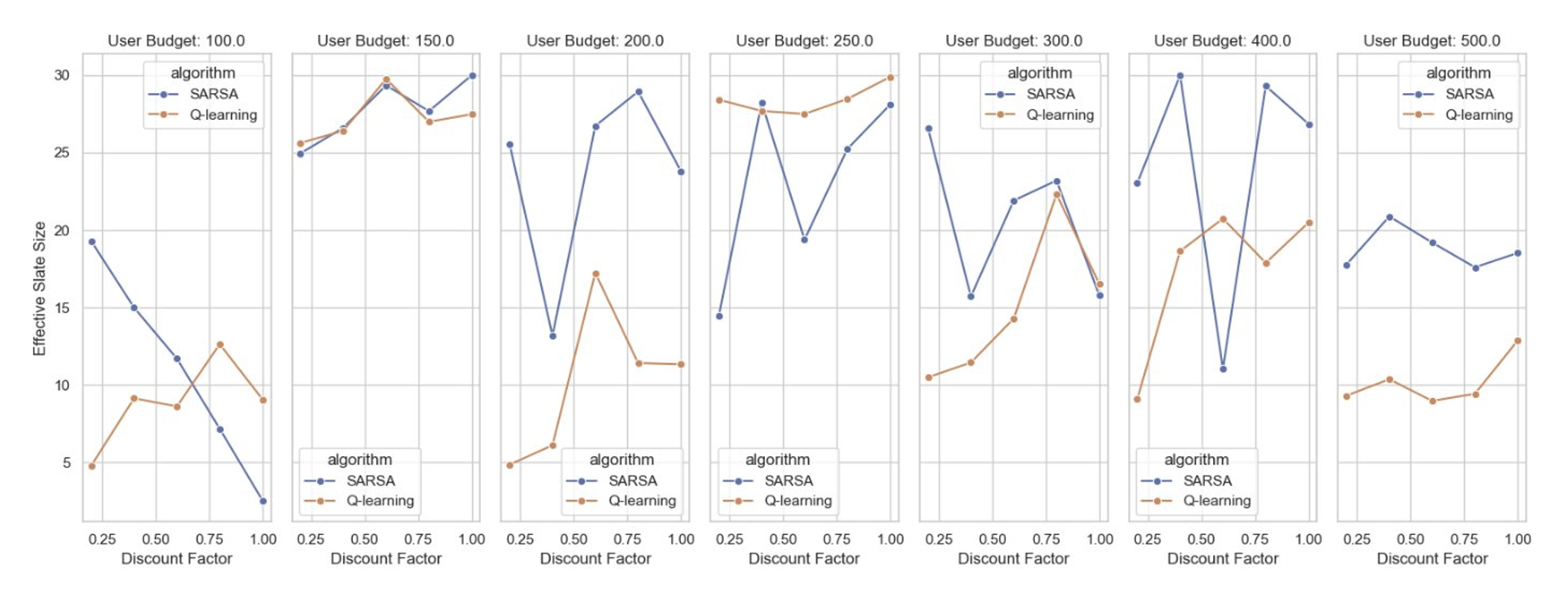}
  \caption{Plots showing variation of Effective Slate Size with discount factor ($\gamma$) with fixed budgets for SARSA \& Q-Learning}
  \label{fig:slate_size}
\end{figure*}

\begin{figure*}[ht]
\centering
  \includegraphics[width=1\linewidth]{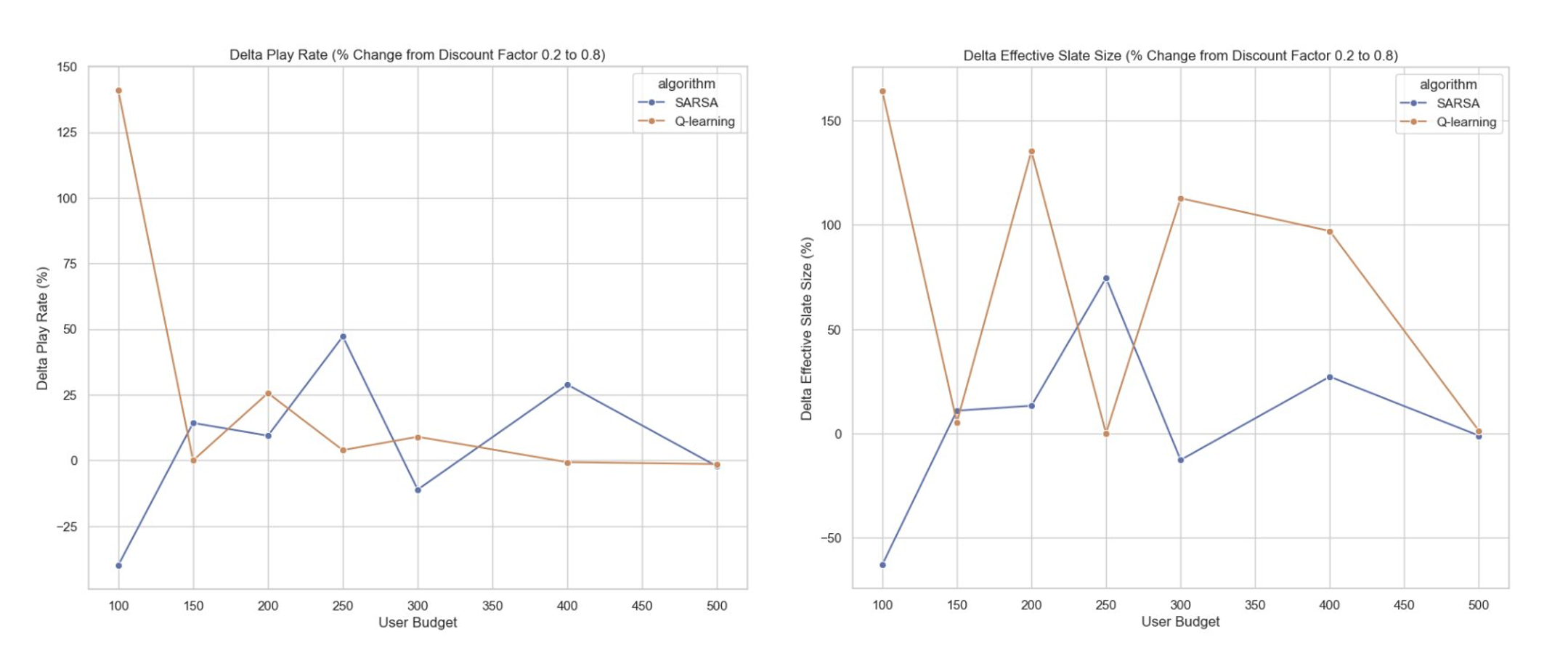}
  \caption{Plots showing variation of delta Play Rate \& delta Effective Slate Size between $\gamma = 0.2$  and $\gamma = 0.8$ with varying user time budgets for both SARSA and Q-Learning}
  \label{fig:delta}
\end{figure*}

Results in Fig.\ref{fig:play_rate} show that while changes in play rate were statistically insignificant, there was a noticeable increase in effective slate size for RL-based models ($\gamma > 0$) compared to contextual bandit models ($\gamma \approx 0$) as seen in Fig.\ref{fig:slate_size}. Comparing the delta play rate between an RL model ($\gamma = 0.2$ vs $\gamma = 0.8$) in Fig.\ref{fig:delta} across different user budget distributions revealed a statistically significant lift in play rate for small to medium budgets. This highlights that RL-based models excel when user budgets are limited and trade-offs between relevance and cost are crucial. Additionally, the RL agent demonstrated superior performance by reducing abandonment probability and increasing effective slate size.

\subsection{Off-Policy Learning}
In realistic settings, where data is generated by a different behavior policy, we evaluated off-policy learning using Q-Learning. Unlike on-policy SARSA, Q-Learning optimizes the value function using Bellman's optimality equation. We compared slates generated by Q-Learning with those from SARSA-trained policies.

Results in Fig.\ref{fig:play_rate} shows minimal change in play rate with respect to discount factor for larger user budgets but produces higher play rate, indicating that Q-Learning achieves better performance to SARSA in terms of user engagement. However, Q-Learning produced significantly larger effective slate sizes as seen in Fig.\ref{fig:slate_size} without a corresponding improvement in play rates. This suggests that while Q-Learning maintains user engagement, it may do so with less efficient slate configurations. Although, initial result are incline toward off-policy, further investigation is necessary to understand the implications of these larger slate sizes and to determine whether they impact other aspects of system performance or user satisfaction.

\section{Conclusion}
In conclusion, this paper presents some preliminary results of the authors' ongoing research showing the comparison of Q-Learning and SARSA techniques used for optimizing slate recommendations.
\begin{itemize}
    \item \textbf{Exploration Effects on Slate Size :} For budget $ u_0 = 100$, Q-Learning performs better than SARSA. Q-Learning's off-policy nature enables it to focus on long-term planning without being hindered by suboptimal exploratory actions, leading to increased effective slate sizes. In contrast, SARSA's on-policy learning incorporates exploratory actions into updates, amplifying their negative impact on future rewards and reducing effective slate size. For the budget in range $100 < u_0 \le 500$, SARSA performs relatively better as the larger time budget reduces the impact of exploratory actions. Longer episodes allow SARSA to recover from suboptimal moves, accumulate rewards, and average out variance introduced by exploration, resulting in more stable updates and improved performance. Notice, Q-Learning improves consistently with higher discount factors due to its ability to prioritize optimal policies without being affected by exploration.

    \item \textbf{Play Rate Trends:} Q-Learning always outperforms SARSA in play rate regardless of the budget. Its robustness to exploration ensures that updates are focused on learning optimal policies, maximizing immediate rewards (e.g., clicks). SARSA, however, is more sensitive to exploration, leading to less accurate value estimates and suboptimal recommendations, which lowers play rate.
    \item \textbf{Time Budget Impact:} While longer episodes help SARSA average out exploration effects, immediate rewards like play rate remain sensitive to suboptimal recommendations. Q-Learning's focus on maximizing expected rewards which gives it a consistent advantage over SARSA across all time budget scenarios.
\end{itemize}
These findings demonstrate that both on-policy and off-policy RL methods are effective for time-constrained recommendation tasks in e-commerce scenarios than traditional methods which involves myopic contextual ranking using contextual bandits ($\gamma = 0$).

\section{Future Work}
Despite the promising results, there are some open questions and possible directions for improvement. Several assumptions especially related to the simulation process, can be relaxed for improving performance allowing more realistic scenarios. 
\begin{enumerate}
    \item \emph{Independent evaluation costs.} Costs $c_i$ are sampled from a simple prior and are position-invariant.  Our aim is to either learn \emph{feature-conditional, position-dependent} costs allowing mixture of distributions or use available real-world dataset with item cost information included as part of the user-item interaction metadata.
    \item \emph{Use policy gradient.} Our current experiments involves only simulation for value based approaches like SARSA and Q-Learning. It would be better to use policy gradient based learning methods in our simulation for improving performance.
    \item \emph{Single-click Bernoulli reward.} We assume at-most one click per slate and $r_t\!\sim\!\mathrm{Bernoulli}(\beta)$. In future, we intend to allow multi-clicks and graded engagement (add-to-cart, purchase) with count or real-valued rewards. By allowing different engagement types, we treat rewards as continuous and calibrated utilities.
    \item \emph{Fixed time budget.} Initial time budgets ($u_0$) are sampled and do not evolve with user satisfaction. Later, our goal is to learn \emph{personalized, time-varying} budgets from behavior by casting the problem as a POMDP with belief over time budget and comparing budget-inference strategies.
\end{enumerate}

\section*{Acknowledgment}

We would like to extend our sincere gratitude to Northwestern University for providing access to their servers and GPU resources, which were instrumental in conducting this research. The computational power and infrastructure made available by the university enabled the efficient processing and analysis of large datasets, significantly contributing to the success of the project. Without this support, the research would not have been possible at the scale or speed required. We deeply appreciate the university’s commitment to fostering a collaborative research environment and supporting technological innovation.

\bibliographystyle{unsrt}  
\bibliography{references}  

\end{document}